\documentclass[utf8]{frontiersSCNS} 

\usepackage{url,lineno,microtype,subcaption}
\usepackage[onehalfspacing]{setspace}
\usepackage{url}

\usepackage{multirow}

\usepackage{rotating}

\usepackage{soul}

\usepackage{float}

\def\keyFont{\fontsize{8}{11}\helveticabold }
\def\firstAuthorLast{Khaki {et~al.}} 
\def\Authors{Saeed Khaki\,$^{1,*}$, Lizhi Wang\,$^{2}$}
\def\Address{$^{1}$Industrial Engineering Department, Iowa State University, Ames, Iowa, USA \\
$^{2}$Industrial Engineering Department, Iowa State University, Ames, Iowa, USA }
\def\corrAuthor{Saeed Khaki}

\def\corrEmail{skhaki@iastate.edu}

\begin{document}
\onecolumn
\firstpage{1}

\title[Crop Yield Prediction Using Deep Neural Networks]{Crop Yield Prediction Using Deep Neural Networks} 

\author[\firstAuthorLast ]{\Authors} 
\address{} 
\correspondance{} 

\extraAuth{}

\maketitle

\begin{abstract}

\section{}
Crop yield is a highly complex trait determined by multiple factors such as genotype, environment, and their interactions. Accurate yield prediction requires fundamental understanding of the functional relationship between yield and these interactive factors, and to reveal such relationship requires both comprehensive datasets and powerful algorithms. In the 2018 Syngenta Crop Challenge, Syngenta released several large datasets that recorded the genotype and yield performances of 2,267 maize hybrids planted in 2,247 locations between 2008 and 2016 and asked participants to predict the yield performance in 2017. As one of the winning teams, we designed a deep neural network (DNN) approach that took advantage of state-of-the-art modeling and solution techniques. Our model was found to have a superior prediction accuracy, with a root-mean-square-error (RMSE) being 12\% of the average yield and 50\% of the standard deviation for the validation dataset using predicted weather data. With perfect weather data, the RMSE would be reduced to 11\% of the average yield and 46\% of the standard deviation.  We also performed feature selection based on the trained DNN model, which successfully decreased the dimension of the input space without significant drop in the prediction accuracy. Our computational results suggested that this model significantly outperformed other popular methods such as Lasso, shallow neural networks (SNN), and regression tree (RT). The results also revealed that environmental factors had a greater effect on the crop yield than genotype. 


\tiny
 \keyFont{ \section{Keywords:} Yield Prediction, Machine Learning, Deep Learning, Feature Selection, Weather Prediction} 
\end{abstract}

\textbf{This paper is published in the Journal of Frontiers in Plant Science\\ available at: \href{ https://doi.org/10.3389/fpls.2019.00621}{ https://doi.org/10.3389/fpls.2019.00621} }

\section{Introduction}

Crop yield prediction is of great importance to global food production. Policy makers rely on accurate predictions to make timely import and export decisions to strengthen national food security \citep{horie1992yield}. Seed companies need to predict the performances of new hybrids in various environments to breed for better varieties \citep{s2}. Growers and farmers also benefit from yield prediction to make informed management and financial decisions \citep{horie1992yield}. However, crop yield prediction is extremely challenging due to numerous complex factors. For example, genotype information is usually represented by high-dimensional marker data, containing many thousands to millions of makers for each plant individual. The effects of the genetic markers need to be estimated, which may be subject to interactions with multiple environmental conditions and field management practices.\\

Many studies have focused on explaining the phenotype (such as yield) as an explicit function of the genotype (G), environment (E), and their interactions (G$\times$E). One of the straightforward and common methods was to consider only additive effects of G and E and treat their interactions as noise \citep{DeLacy1996,Heslot2014}. A popular approach to study the G$\times$E effect was to identify the effects and interactions of mega environments rather than more detailed environmental components. Several studies proposed to cluster the environments based on discovered drivers of G$\times$E interactions \citep{Chapman2000,Cooper1994}. \cite{crossa1997sites,crossa1995shifted} used the sites regression and the shifted multiplicative models for G$\times$E interaction analysis by dividing environments into similar groups. \cite{burgueno2008using} proposed an integrated approach of factor analytic (FA) and linear mixed models to cluster environments and genotypes and detect their interactions. They also stated that FA model can improve predictability up to 6\% when there were complex G$\times$E patterns in the data \citep{burgueno2011prediction}. Linear mixed models have also been used to study both additive and interactive effects of individual genes and environments \citep{crossa2004studying, Montesinos-Lopez2018}.

More recently, machine learning techniques have been applied for crop yield prediction, including multivariate regression, decision tree, association rule mining, and artificial neural networks. A salient feature of machine learning models is that they treat the output (crop yield) as an implicit function of the input variables (genes and environmental components), which could be a highly non-linear and complex function. \cite{Liu2001} employed a neural network with one hidden layer to predict corn yield using input data on soil, weather, and management. \cite{Drummond2003} used stepwise multiple linear regression, projection pursuit regression, and neural networks to predict crop yield, and they found that their neural network model outperformed the other two methods. \cite{marko2016soybean} proposed weighted histograms regression to predict the yield of different soybean varieties, which demonstrated superior performances over conventional regression algorithms. \cite{romero2013using} applied decision tree and association rule mining to classify yield components of durum wheat.

In this paper, we use deep neural networks to predict yield, check yield, and yield difference of corn hybrids from genotype and environment data. Deep neural networks belong to the class of representation learning models that can find the underlying representation of data without handcrafted input of features. Deep neural networks have multiple stacked non-linear layers which transform the raw input data into higher and more abstract representation at each stacked layer \citep{LeCun2015}. As such, as the network grows deeper, more complex features are extracted which contribute to the higher accuracy of results. Given the right parameters, deep neural networks are known to be universal approximator functions, which means that they can approximate almost any function, although it may be very challenging to find the right parameters \citep{Goodfellow2016,Hornik1990}.

Compared with the aforementioned neural network models in the literature, which were shallow networks with a single hidden layer, deep neural networks with multiple hidden layers are more powerful to reveal the fundamental non-linear relationship between input and response variables \citep{LeCun2015}, but they also require more advanced hardware and optimization techniques to train. For example, the neural network's depth (number of hidden layers) has significant impact on its performance. Increasing the number of hidden layers may reduce the classification or regression errors, but it may also cause the vanishing/exploding gradients problem that prevents the convergence of the neural networks \citep{He2016,Glorot2010,Bengio1994}. Moreover, the loss function of the deep neural networks is highly non-convex due to having numerous non-linear activation functions in the network. As a result, there is no guarantee on the convergence of any gradient based optimization algorithm applied on neural networks \citep{Goodfellow2016}. There have been many attempts to solve the gradient vanishing problem, including normalization of the input data, batch normalization technique in intermediate layers, stochastic gradient descent (SGD) \citep{LeCun1998,Ioffe2015}, and using multiple loss functions for intermediate layers \citep{Szegedy2015}. However, none of these approaches would be effective for very deep networks. \cite{He2016} argued that the biggest challenge with deep neural networks was not overfitting, which can be addressed by adding regularization or dropout to the network \citep{Srivastava2014}, but it was the structure of the network. They proposed a new structure for deep neural networks using identity blocks or residual shortcuts to make the optimization of deeper networks easier \citep{He2016}. These residual shortcuts act like a gradient highway throughout the network and prevent vanishing gradient problem.

Deep learning models have recently been used for crop yield prediction. \cite{you2017deep} used deep learning techniques such as convolutional neural networks and recurrent neural networks to predict soybean yield in the United States based on  a sequence of remotely sensed images taken before the
harvest. Their model outperformed traditional remote-sensing based methods by 15\%  in terms of Mean Absolute Percentage Error (MAPE). \cite{russello2018convolutional} used convolutional neural networks for crop yield prediction based on satellite images. Their model used 3-dimensional convolution to include spatiotemporal features, and outperformed other machine learning methods.

The remainder of this paper is organized as follows. Section 2 describes the data used in this research. Section 3 provides a detailed description of our deep neural networks for yield prediction. Section 4 presents the results of our model. Section 5 describes the feature selection method. Finally, we conclude the paper in section 6.

\section{Data}

\begin{figure}[h]
\begin{center}
\includegraphics[scale=0.23]{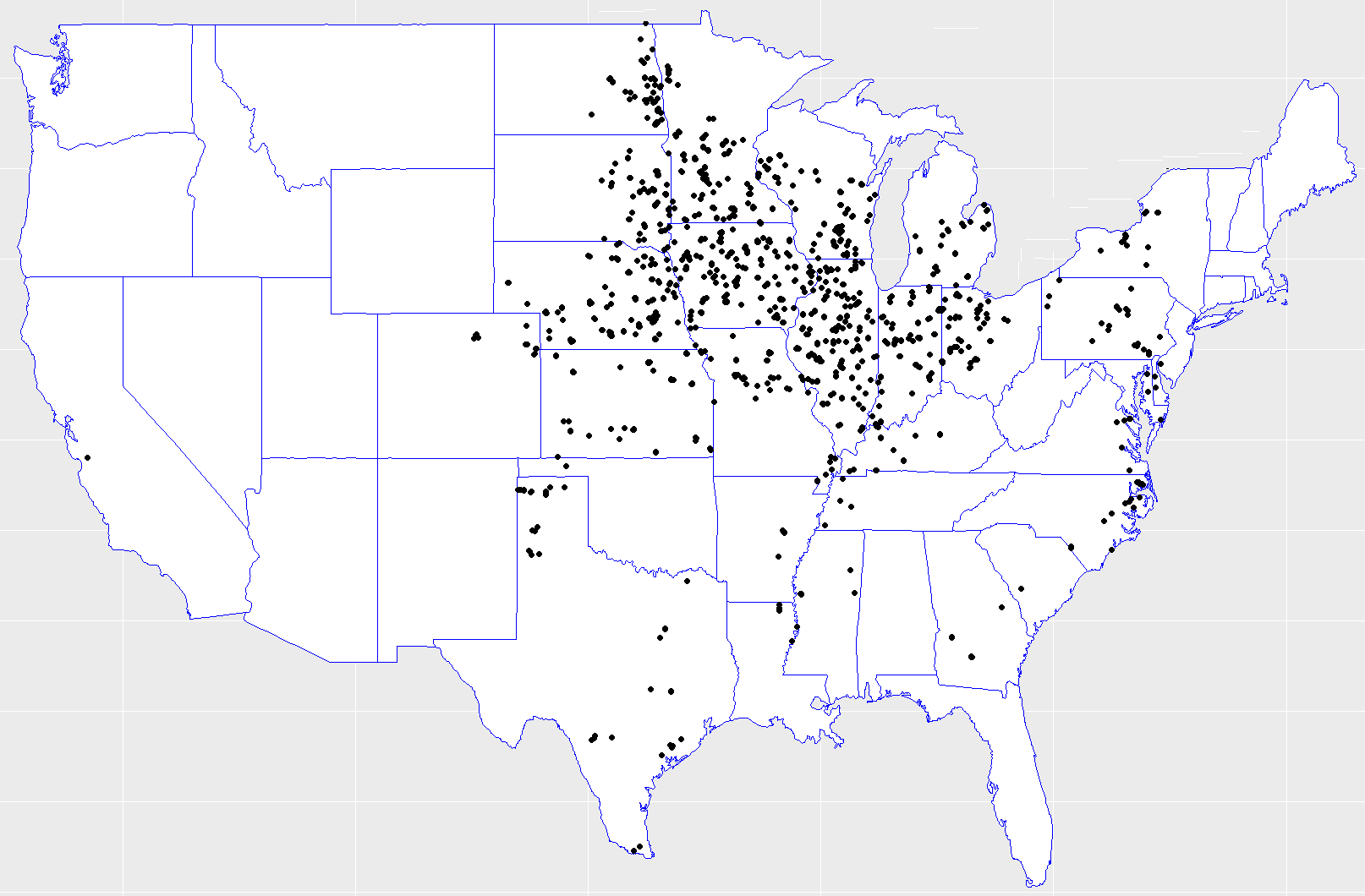}
\end{center}
\caption{Hybrids locations across the United States. Data collected from the 2018 Syngenta Crop Challenge \cite{s2}.}
\label{fig:location}
\end{figure}

In the 2018 Syngenta Crop Challenge \citep{s2}, participants were asked to use real-world data to predict the performance of corn hybrids in 2017 in different locations. The dataset included 2,267 experimental hybrids planted in 2,247 of locations between 2008 and 2016 across the United States and Canada. Most of the locations were across the United States. This was one of the largest and most comprehensive datasets that were publicly available for research in yield prediction, which enabled the deployment and validation of the proposed deep neural network model. Figure \ref{fig:location} shows the distribution of hybrids across the United States.

The training data included three sets: crop genotype, yield performance, and environment (weather and soil). The genotype dataset contained genetic information for all experimental hybrids, each having 19,465 genetic markers. The yield performance dataset contained the observed yield, check yield (average yield across all hybrids of the same location), and yield difference of 148,452 samples for different hybrids planted in different years and locations. Yield difference is the difference between yield and check yield, and indicates the relative performance of a hybrid against other hybrids at the same location \citep{marko2017portfolio}. The environment dataset contained 8 soil variables and 72 weather variables (6 weather variables measured for 12 months of each year). The soil variables included percentage of clay, silt and sand, available water capacity (AWC), soil pH, organic matter (OM), cation-exchange capacity (CEC), and soil saturated hydraulic conductivity (KSAT). The weather data provided in the 2018 Syngenta Crop Challenge were normalized and anonymized. Based on the pattern of the data, we hypothesized that they included day length, precipitation, solar radiation, vapor pressure, maximum temperature, and minimum temperature.


The goal of the 2018 Syngenta Crop Challenge was to predict the performance of corns in 2017, but the ground truth response variables for 2017 were not released after the competition. In this paper, we used the 2001 to 2015 data and part of the 2016 data as the training dataset (containing 142,952 samples) and the remaining part of the 2016 data as the validation dataset (containing 5,510 samples). All validation samples were unique combinations of hybrids and locations, which did not have any overlap with training data.

\section{Methodology}

\subsection{Data Preprocessing}

The genotype data were coded in $\{-1,0,1\}$ values, respectively representing aa, aA, and AA alleles.
Approximately 37\% of the genotype data had missing values. To address this issue, we used a two-step approach to preprocess the genotype data before they can be used by the neural network model. First, we used a 97\% call rate to discard genetic markers whose non-missing values were below this call rate. Then we also discarded genetic markers whose lowest frequent allele's frequency were below 1\%, since these markers were less heterozygous and therefore less informative. As a result, we reduced the number of genetic markers from 19,465 to 627. To impute the missing data in the remaining part of the genotype data, we tried multiple imputation techniques, including mean, median, and most frequent \citep{allison2001missing}, and found that the median approach led to the most accurate predictions. The yield and environment datasets were complete and did not have missing data.

\subsection{Weather Prediction}

Weather prediction is an inevitable part of crop yield prediction, because weather plays an important role in yield prediction but it is unknown a priori. In this section, we describe our approach for weather prediction and apply it to predict the 2016 weather variables using the 2001-2015 weather data.

Let $X_{l,y}^w$ denote the weather variable $w$ at location $l$ in year $y$, for all $w \in \{1, ..., 72\}$, $l \in \{1, ..., 2247\}$, and $y \in \{2001, ..., 2016\}$. To predict the 2016 weather variables using historical data from 2001 to 2015, we trained 72 shallow neural networks for the 72 weather variables, which were used across all locations. There were two reasons for the aggregation of 2,247 locations: (1) the majority of the locations were in the middle west region, so it was reasonable to make the simplifying assumption that the prediction models were uniform across locations, (2) combining historical data for all locations allows sufficient data to train the 72 neural networks more accurately.

\begin{figure}[H]
\begin{center}
\begin{picture}(0,140)(100,0)
\put(25,0){\includegraphics[width=3.0in]{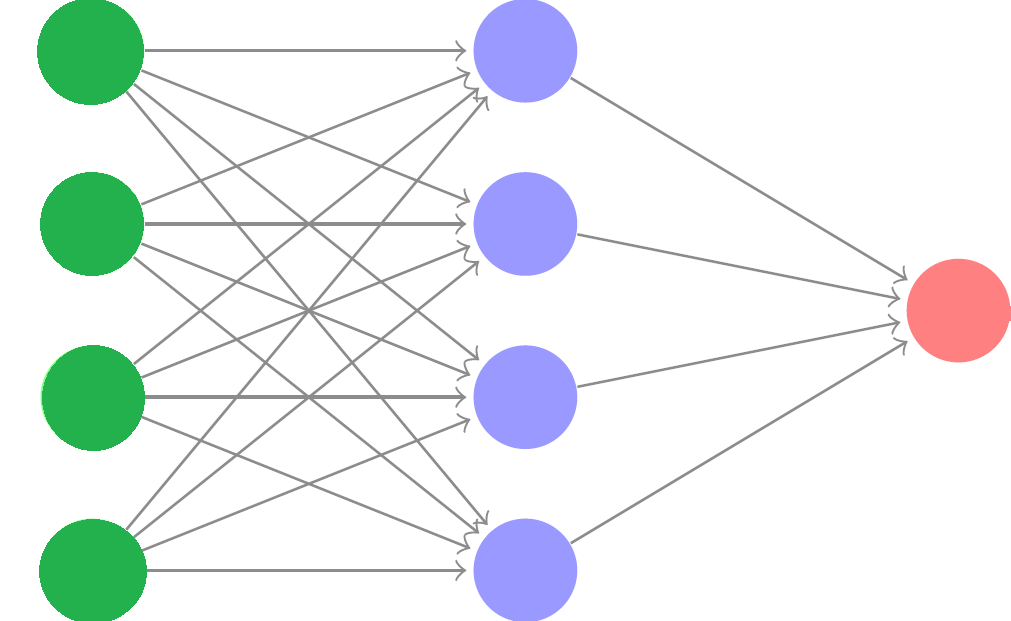}}
\put(250,64){$X_{l,y}^w$}
\put(0,120){$X_{l,y-1}^w$}
\put(0,85){$X_{l,y-2}^w$}
\put(0,50){$X_{l,y-3}^w$}
\put(0,10){$X_{l,y-4}^w$}
\end{picture}
\end{center}
\caption{Neural network structure for weather prediction with a 4-year lag.}
\label{fig:nnw}
\end{figure}
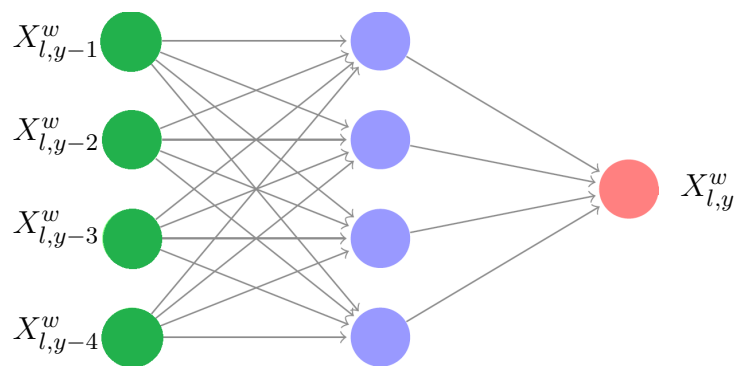

For each weather variable $w$, the neural network model explains the weather variable $X_{l,y}^w$ at location $l$ in year $y$ as a response of four previous years at the same location: $\{X_{l,y-1}^w,X_{l,y-2}^w,X_{l,y-3}^w,X_{l,y-4}^w\}$. We have tried other parameters for the periodic lag and found four years to yield the best results. As such, there were $24,717$ samples of training data for each weather variable. The resulting parameters of the networks were then used to predict $X_{l,y=2016}^w$ using historical data of $X_{l,y=2012}^w$ to $X_{l,y=2015}^w$ for all $l$ and $w$. The structure of a shallow neural network is given in Figure \ref{fig:nnw}.

The reason for using neural networks for weather prediction is that neural networks can capture the nonlinearities, which exist in the nature of weather data, and they learn these nonlinearities from data without requiring the nonlinear model to be specified before estimation \citep{abhishek2012weather}. Similar neural network approaches have also been used for other weather prediction studies \citep{abhishek2012weather,maqsood2004ensemble,
bou2017using,kaur2011artificial,bustami2007artificial,
baboo2010efficient}.

\subsection{Yield Prediction Using Deep Neural Networks}

\begin{figure}[h]
\begin{center}

\includegraphics[scale=0.45]{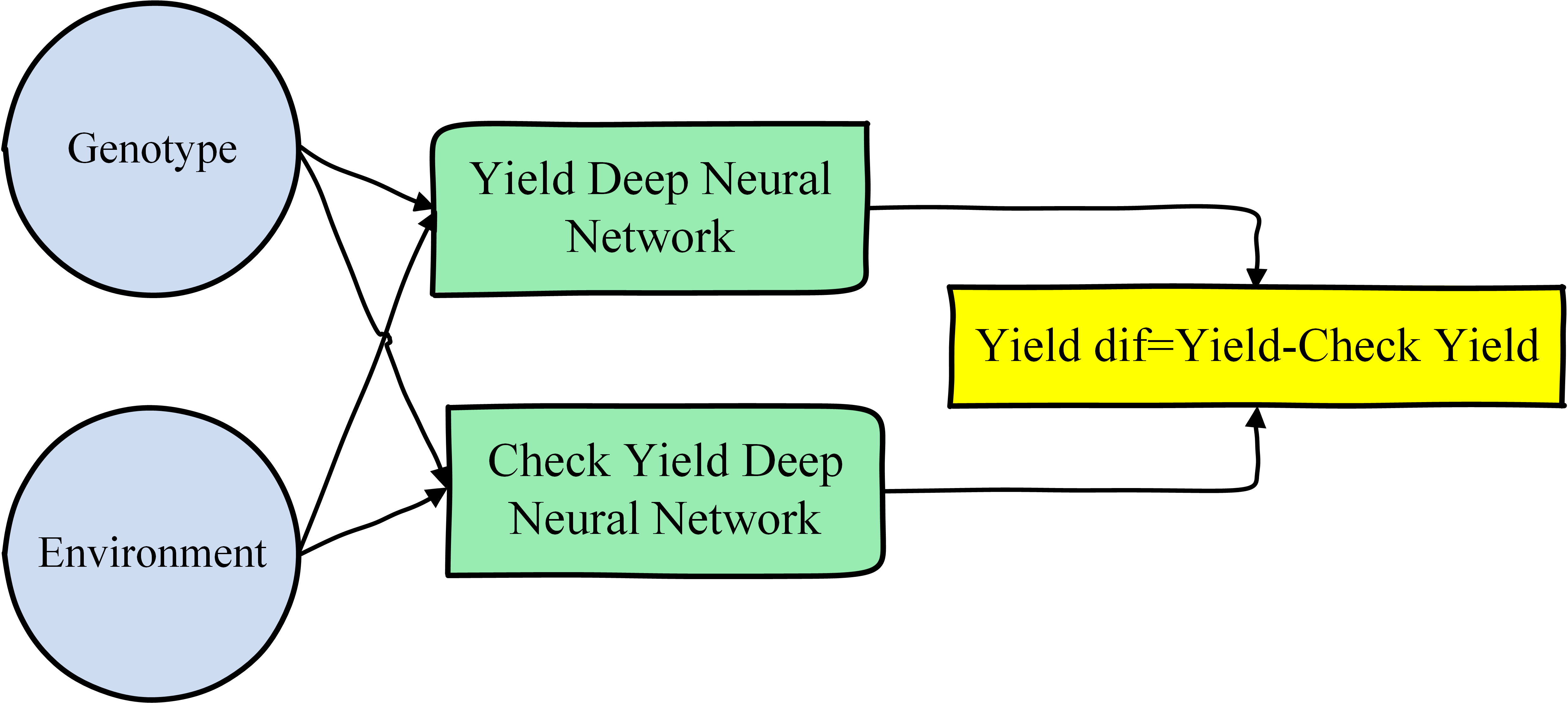}
\end{center}
\caption{Neural networks designed for predicting yield difference.}
\label{fig:nd}
\end{figure}

We trained two deep neural networks, one for yield and the other for check yield, and then used the difference of their outputs as the prediction for yield difference. These models are illustrated in Figure \ref{fig:nd}. This model structure was found to be more effective than using one single neural network for yield difference, because the genotype and environment effects are more directly related to the yield and check yield than their difference.


\begin{figure}[H]

\begin{center}

\includegraphics[width=\linewidth]{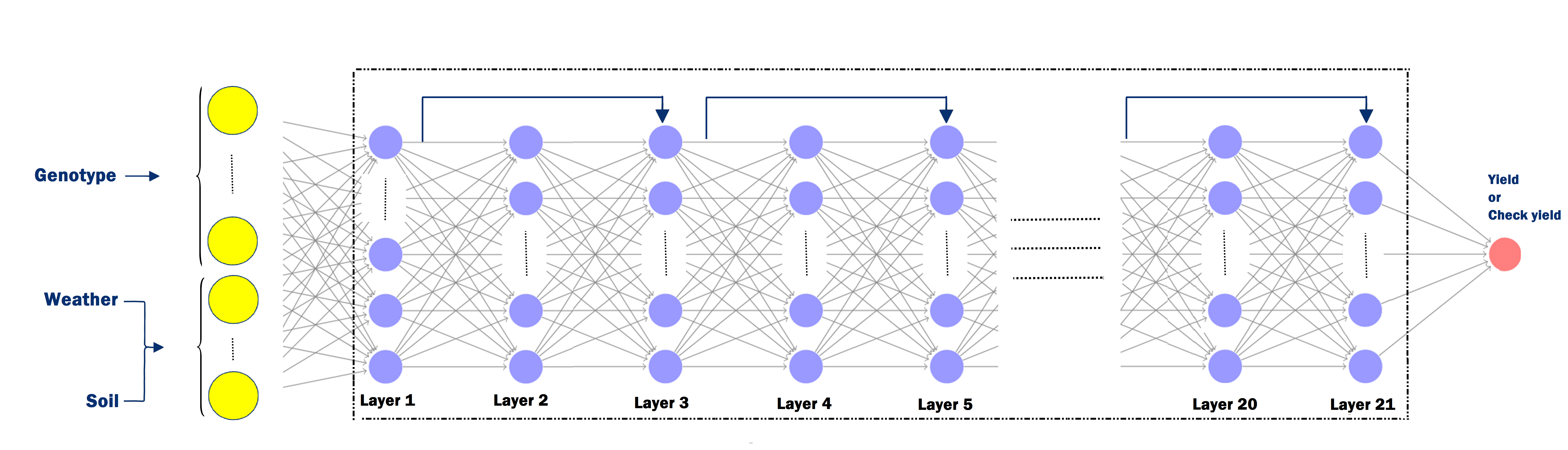}

\end{center}
\caption{Deep neural network structure for yield or check yield prediction. The input layer takes in genotype data ($G \in \mathbb{Z}^{n\times p}$), weather data ($W \in \mathbb{R}^{n\times k_1}$), and soil data ($S \in \mathbb{R}^{n\times k_2}$) as input. Here, $n$ is the the number of observations, $p$ is the number of genetic markers, $k_1$ is the number of weather components, and $k_2$ is the number of soil conditions. Odd numbered layers have a residual shortcut connection which skips one layer. Each sample is fed to the network as a vector with dimension of $\mathbb{R}^{ \,p+k_1+k_2}$.}

\label{fig:aa1}
\end{figure}

The following hyperparameters were used in the training process. Each neural network has 21 hidden layers and 50 neurons in each layer. After trying deeper network structures, these dimensions were found to provide the best balance between prediction accuracy and limited overfitting. We initialized all weights with the Xavier initialization method \citep{Glorot2010}. We used SGD with a mini-batch size of 64. The Adam optimizer was used with a learning rate of 0.03\%, which was divided by 2 every 50,000 iterations \citep{Kingma2014}. Batch normalization was used before activation for all hidden layers except the first hidden layer. Models were trained for 300,000 maximum iterations. Residual shortcuts were used for every two stacked hidden layers \citep{He2016}. We used maxout activation \citep{Goodfellow2016} function \cite{} for all neurons in the networks except for the output layer, which did not have any activation function. In order to avoid overfitting, we used the $L_2$ regularization \citep{ng2004feature} for all hidden layers. We also added $L_1$ regularization \citep{ng2004feature} to the first layer to decrease the effect of redundant features, as in Lasso \citep{tibshirani1996regression}. Figure \ref{fig:aa1} depicts the detailed structure of the deep neural network, which was the same for yield and check yield prediction.

\section{Results}

The two deep neural networks were implemented in Python using the Tensorflow open-source software library \citep{Abadi2016}. The training process took approximately 1.4 hours on a Tesla K20m GPU for each neural network. We also implemented three other popular prediction models for comparison: The least absolute shrinkage and selection operator (Lasso), shallow neural network (having a single hidden layer with 300 neurons), and regression tree \citep{breiman2017classification}. To ensure fair comparisons, two sets of these three models were built to predict yield and check yield separately, and the differences of their outputs were used as the prediction for the yield difference. All of these models were implemented in Python in the most efficient manner that we were capable of and tested under the same software and hardware environments to ensure fair comparisons.

The following hyperparameters were used for the regression tree.  The maximum depth of the tree was set to 10 to avoid overfitting. We set the minimum number of samples required to split an internal node of tree to be 2. All features were used to train the regression tree. We tried different values for the coefficient of $L_1$ term \citep{ng2004feature} in the Lasso model, and found that values between 0.1 and 0.3 led to the most accurate predictions.

\setlength{\extrarowheight}{10pt}
\begin{table}[H]
\begin{center}

  \begin{scriptsize}
  \begin{adjustwidth}{0cm}{}
  \centering
  \begin{tabular}{|c|c|c|c|c|c|}
    \hline
\begin{turn}{-90}Model\end{turn}&{\scriptsize \begin{tabular}[t]{@{}c@{}}Response\\Variable\end{tabular}} &{\scriptsize \begin{tabular}[t]{@{}c@{}}Training\\RMSE\end{tabular}} &{\scriptsize \begin{tabular}[t]{@{}c@{}}Training\\Correlation\\  Coefficient (\%)\end{tabular}}& {\scriptsize \begin{tabular}[t]{@{}c@{}}Validation\\RMSE\end{tabular}}& {\scriptsize \begin{tabular}[t]{@{}c@{}}Validation\\Correlation\\  Coefficient (\%)\end{tabular}} \\
\hline
 \multirow{3}{*}{\begin{turn}{-90}{DNN}\end{turn}}&{Yield} & 10.55 & 88.3 & 12.79 & 81.91  \\[7pt]
    \cline{2-6}

  &{\scriptsize Check yield} &8.21 & 91.00 & 11.38& 85.46 \\
    \cline{2-6}
   &{Yield difference} &  11.79 & 45.87 &  12.40 &29.28 \\
    \cline{1-6}

\multirow{3}{*}{\begin{turn}{-90}{Lasso}\end{turn}}&{Yield} & 20.28 & 36.68  & 21.40 & 27.56  \\[7pt]
    \cline{2-6}

  &{\scriptsize Check yield} &18.85 & 28.49 & 19.87& 23.00 \\
    \cline{2-6}
   &{Yield difference} &  15.32 & 19.78& 13.11 &6.84         \\
    \cline{1-6}

\multirow{3}{*}{\begin{turn}{-90}{SNN}\end{turn}}&{Yield} & 12.96 & 80.21  &  18.04 &60.11 \\[7pt]
    \cline{2-6}

  &{\scriptsize Check yield} & 10.24 &71.18 & 15.18 &60.48 \\
    \cline{2-6}
   &{Yield difference} &  9.92 &58.74& 15.19 & 11.39            \\
    \cline{1-6}

\multirow{3}{*}{\begin{turn}{-90}{RT}\end{turn}}&{Yield} & 14.31 & 76.7 & 15.03 & 73.8  \\[7pt]
    \cline{2-6}

  &{\scriptsize Check yield} &14.55 & 82.00 & 14.87& 69.95 \\
    \cline{2-6}
   &{Yield difference} &  17.62 & 21.12 & 15.92 &5.1            \\
    \cline{1-6}

  \end{tabular}

 \caption{Prediction performance with ground truth weather variables. DNN, Lasso, SNN, and RT stand for deep neural networks, least absolute shrinkage and selection operator, shallow neural network, and regression tree, respectively. The average$\pm$standard deviation for yield, check yield, and yield difference are, respectively, 116.51$\pm$27.7, 128.27$\pm$25.34, and $-$11.76$\pm$ 14.27.}\label{tab2}
  \end{adjustwidth}
\end{scriptsize}
\end{center}
\end{table}

Table \ref{tab2} compares the performances of the four models on both training and validation datasets with respect to the RMSE and correlation coefficient. These results suggest that the deep neural networks outperformed the other three models to varying extents. The weak performance of Lasso was mainly due to its linear model structure, which ignored epistatic or G$\times$E interactions and the apparent nonlinear effects of environmental variables. SNN outperformed Lasso on all the performance measures except validation RMSE of the yield difference, since it was able to capture nonlinear effects. As a non-parametric model, RT demonstrated comparable performance with SNN with respect to yield and check yield but was much worse with respect to the yield difference. DNN outperformed all of the three benchmark models with respect to almost all measures; the only exception was that SNN had a better performance for the training dataset but worse for the validation dataset, which was a sign of overfitting. The DNN model was particularly effective in predicting yield and check yield, with RMSE for the validation dataset being approximately 11\% of their respective average values. The accuracy for the check yield was a little higher than that for the yield because the former is the average yield across all hybrids and all years for the same location, which is easier to predict than the yield for individual hybrid at a specific location in a specific year. The model struggled to achieve the same prediction accuracy for yield difference as for the other two measures, although it was still significantly better than the other three benchmark models. The Lasso's performance seemed good for the yield difference with respect to RMSE, but it had a low correlation coefficient. This happened because Lasso's prediction was much centralized around the mean which may increase the risk of getting high prediction error on other test data. Let $y$, $y_c$, and $y_d$ denote yield, check yield, and yield difference, respectively. Then, the variance of yield difference can be defined as Equation \ref{eq:yd}. As shown in Equation \ref{eq:yd}, the yield difference was more difficult to predict since its variation depends on not only the individual variances of yield and check yield but also their covariance.
\begin{align}
Var(y_d)=Var(y-y_c)=Var(y)+Var(y_c)-2\,Cov(y,y_c)\label{eq:yd}
\end{align} 

\begin{figure}[H]
\begin{center}
\includegraphics[scale=0.3]{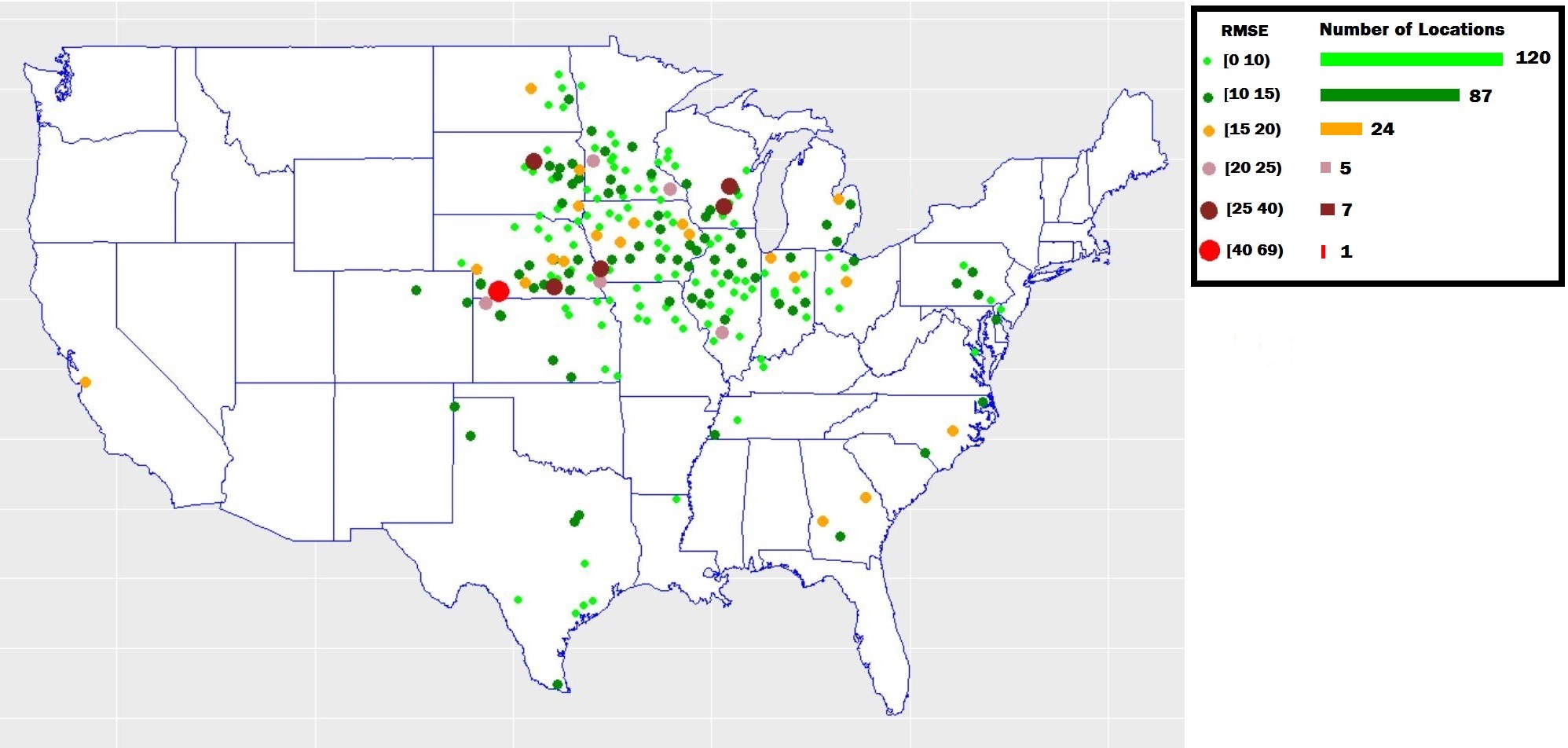}
\end{center}
\caption{The yield prediction error for individual regions in the validation dataset. The map shows the validation locations across the United States.}
\label{fig:location_rmse}
\end{figure}

\begin{figure}[H]
\begin{center}
\includegraphics[scale=0.30]{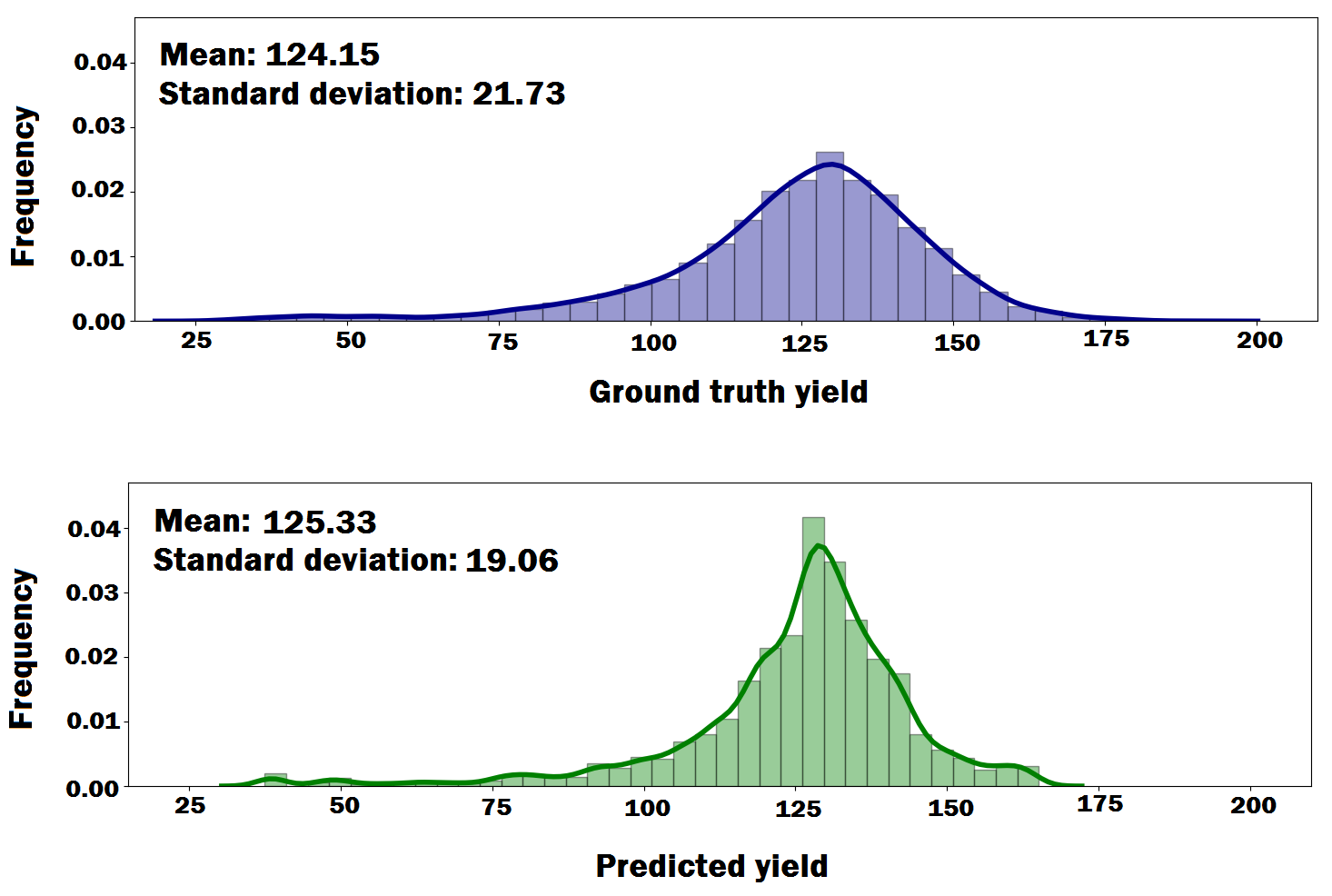}
\end{center}
\caption{The probability density functions of the ground truth yield and the predicted yield by DNN model. The plots indicate that DNN model can approximately preserve the distributional properties of the ground truth yield.}
\label{fig:dist}
\end{figure}

To examine the yield prediction error for individual regions, we obtained prediction error across 244 locations existed in the validation dataset. As shown in Figure \ref{fig:location_rmse}, the prediction error was consistently low (RMSE below 15) for most of locations (207 locations).
We plotted the probability density functions of the ground truth yield and the predicted yield by the DNN model to see if the DNN model can preserve the distributional properties of the ground truth yield. As shown in Figure \ref{fig:dist}, the DNN model can approximately preserve the distributional properties of the ground truth yield. However, the variance of the predicted yield is less than the variance of the ground truth yield, which indicates DNN model's prediction was more centralized around mean.

To evaluate the effects of weather prediction on the performance of the DNN model, we obtained prediction results using the predicted weather data rather than the ground truth weather data. As shown in Table \ref{tab3}, the prediction accuracy of DNN deteriorated compared to the corresponding results in Table \ref{tab2}, which suggested how sensitive yield prediction is to weather prediction and the extent to which a perfect weather prediction model would improve the yield prediction results.

\setlength{\extrarowheight}{10pt}
\begin{table}[H]
\begin{scriptsize}
\begin{adjustwidth}{-0.0cm}{}
  \centering
 \begin{tabular}{|c|c|c|c|c|c|}
    \hline
\begin{turn}{-90}Model\end{turn}&{\scriptsize \begin{tabular}[t]{@{}c@{}}Response\\Variable\end{tabular}} &{\scriptsize \begin{tabular}[t]{@{}c@{}}Training\\RMSE\end{tabular}} &{\scriptsize \begin{tabular}[t]{@{}c@{}}Training\\Correlation\\  Coefficient (\%)\end{tabular}}& {\scriptsize \begin{tabular}[t]{@{}c@{}}Validation\\RMSE\end{tabular}}& {\scriptsize \begin{tabular}[t]{@{}c@{}}Validation\\Correlation\\  Coefficient (\%)\end{tabular}} \\
\hline
 \multirow{3}{*}{\begin{turn}{-90}{\scriptsize DNN}\end{turn}}&{Yield} & 11.64& 85.66 & 13.94& 78.65   \\[7pt]
    \cline{2-6}

  &{\scriptsize Check yield} & 9.49 & 78.35 & 12.51 &75.04 \\
    \cline{2-6}
   &{Yield difference} & 12.80 &37.64 & 15.54& 19.86        \\
    \cline{1-6}

  \end{tabular}
 \caption{Prediction performance with predicted weather variables.}\label{tab3}
  \end{adjustwidth}

\end{scriptsize}
\end{table}

\section{Analysis}

\subsection{Importance Comparison Between Genotype and Environment}

To compare the individual importance of genotype, soil and weather components in the yield prediction, we obtained the yield prediction results using following models:

\textbf{DNN(G)}: This model uses the DNN model to predict the phenotype based on the genotype data (without using the environment data), which is able to capture linear and nonlinear effects of genetic markers. 

\textbf{DNN(S)}: This model uses the DNN model to predict the phenotype based on the soil data (without using the genotype and weather data), which is able to capture linear and nonlinear effects of soil conditions. 

\textbf{DNN(W)}: This model uses the DNN model to predict the phenotype based on the weather data (without using the genotype and soil data), which is able to capture linear and nonlinear effects of weather components.

\textbf{Average}: This model provides a baseline using only the average of phenotype for prediction.

Table \ref{tab4} compares the performances of the above 4 models in the yield prediction. The results suggested that DNN(W) and DNN(S) had approximately the same performance and their prediction accuracies were significantly higher than DNN(G), which revealed that the environmental (weather and soil) components explained more of the variation within the crop yield compared to genotype.

\setlength{\extrarowheight}{10pt}
\begin{table}[H]
\begin{scriptsize}
\begin{adjustwidth}{-0.0cm}{}
  \centering
 \begin{tabular}{|c|c|c|c|c|}
    \hline
{\scriptsize \begin{tabular}[t]{@{}c@{}}\\Model\end{tabular}} &{\scriptsize \begin{tabular}[t]{@{}c@{}}Training\\RMSE\end{tabular}} &{\scriptsize \begin{tabular}[t]{@{}c@{}}Training\\Correlation\\  Coefficient (\%)\end{tabular}}& {\scriptsize \begin{tabular}[t]{@{}c@{}}Validation\\RMSE\end{tabular}}& {\scriptsize \begin{tabular}[t]{@{}c@{}}Validation\\Correlation\\  Coefficient (\%)\end{tabular}} \\
\hline
 {DNN(G)} & 21.74& 20.26 & 21.72& 15.09   \\[7pt]
    \hline

  {DNN(S)} & 15.28 & 73.37 & 15.49 &72.04 \\
    \hline
   {DNN(W)} & 14.26 &76.98 & 14.96& 72.60        \\
    \hline
    {Average} & 24.40 &0.0 & 23.14& 0.0        \\
    \hline

  \end{tabular}
 \caption{Yield prediction performances of DNN(G), DNN(S), DNN(W), and Average model.}\label{tab4}
  \end{adjustwidth}

\end{scriptsize}
\end{table}

\newpage
\subsection{ Feature Selection}

Genotype and environment data are often represented by many variables, which do not have equal effect or importance in yield prediction. As such, it is vital to find important variables and omit the other redundant ones which may decrease the accuracy of predictive models. In this paper, we used guided backpropagation method which backpropagates the positive gradients to find input variables which maximize the activation of our interested neurons \citep{springenberg2014striving}. As such, it is not important if an input variable suppresses a neuron with negative gradient somewhere along the path to our interested neurons.

\begin{figure}[H]
\begin{center}
\includegraphics[height=4cm,width=\linewidth]{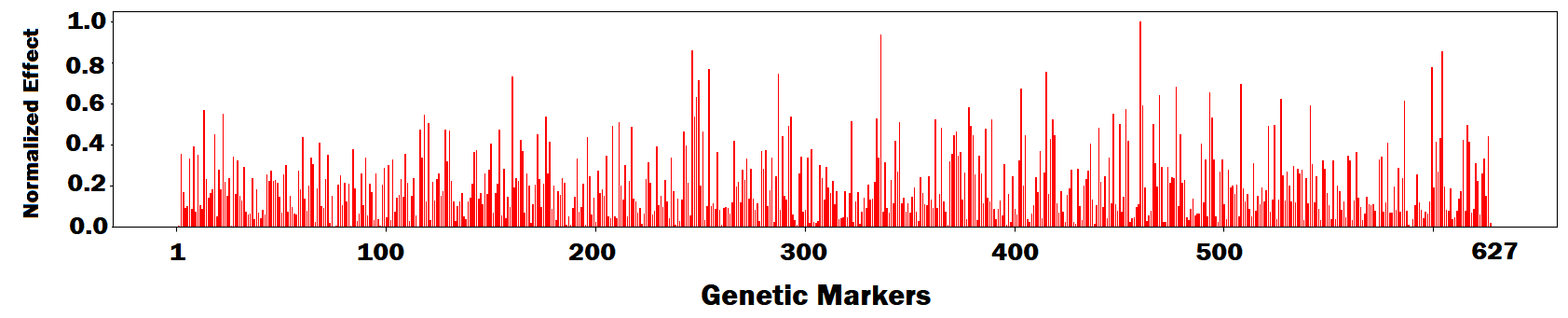}
\end{center}
\caption{Bar plot of estimated effects of 627 genetic markers.}
\label{fig:g_grad}
\end{figure}

\begin{figure}[H]
\begin{center}
\includegraphics[scale=0.15]{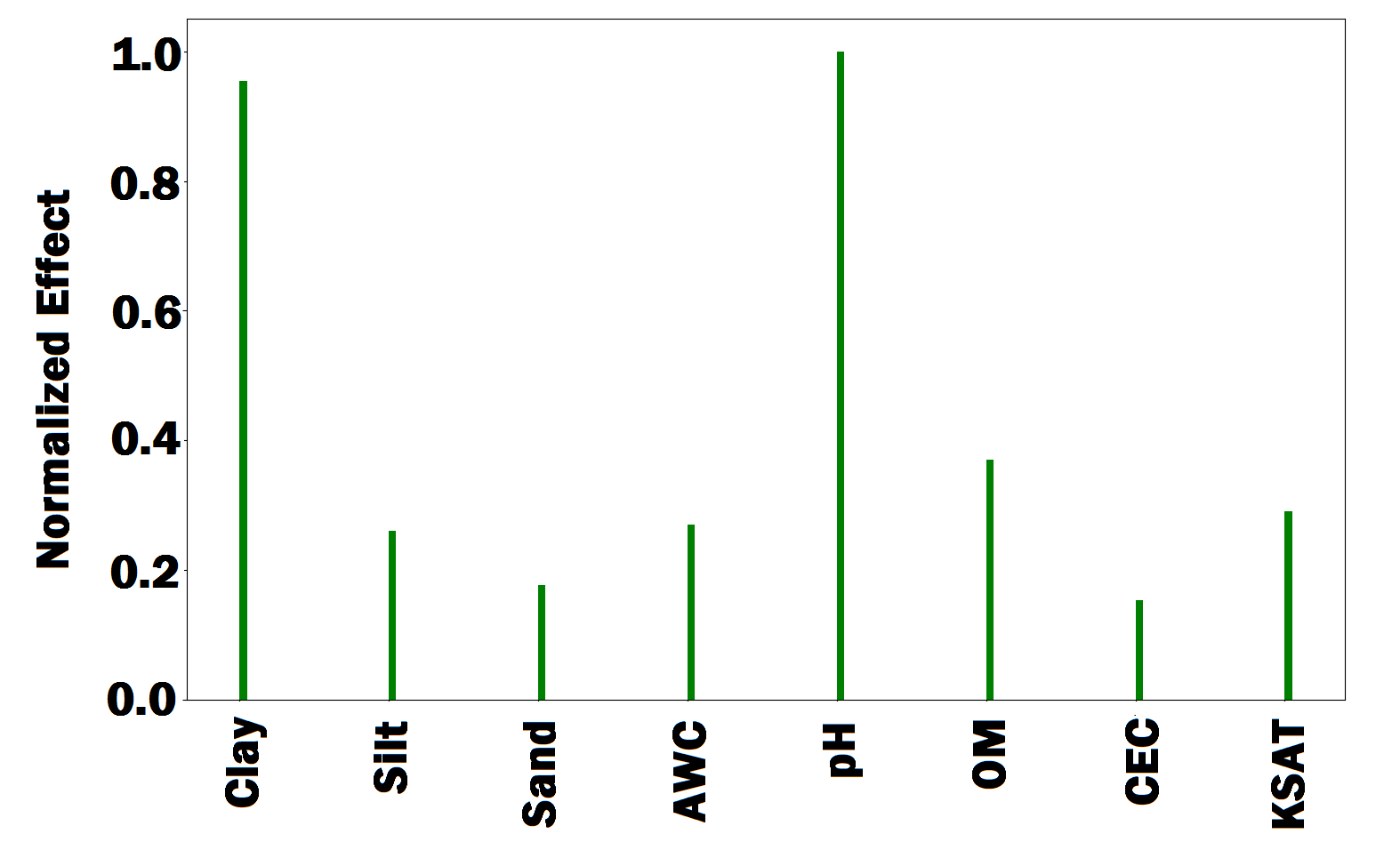}
\end{center}
\caption{Bar plot of estimated effects of 8 soil conditions. AWC, OM, CEC, and KSAT stand for available water capacity, organic matter, cation exchange capacity, and saturated hydraulic conductivity, respectively. The Bar plot indicates that percentage of clay and soil pH were more important than the other soil conditions.}
\label{fig:s_grad}
\end{figure}

\begin{figure}[H]
\begin{center}
\includegraphics[scale=0.15]{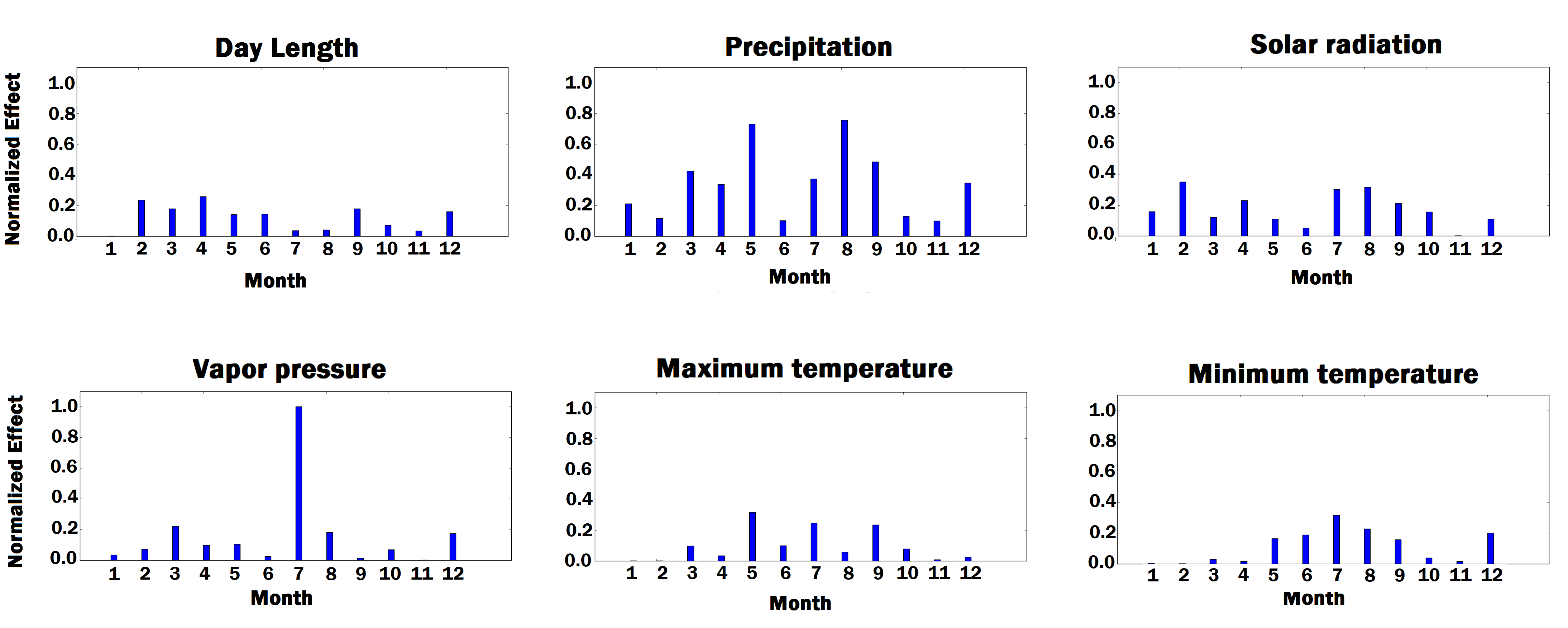}
\end{center}
\caption{Bar plot of estimated effects of 6 weather components measured for 12 months of each year, starting from January. The vertical axes were normalized across all weather components to make the effects comparable.}
\label{fig:w_grad}
\end{figure}

First, we fed all validation samples to the DNN model and computed the average activation of all neurons in the last hidden layer of the network. We set the gradient of activated neurons to be 1 and the other neurons to be 0. Then, the gradients of the activated neurons were backpropagated to the input space to find the associated input variables based on the magnitude of the gradient (the bigger, the more important). Figures \ref{fig:g_grad}, \ref{fig:s_grad}, and \ref{fig:w_grad} illustrate the estimated effects of genetic markers, soil conditions, and weather components, respectively. The estimated effects indicate the relative importance of each feature compared to the other features. The effects were normalized within each group namely, genetic markers, soil conditions, and weather components to make the effects comparable.

As shown in Figure \ref{fig:w_grad}, solar radiation and temperature have considerable effects on the variation in corn yield across different environments. High corn yield is associated with low temperature and high solar radiation since lower temperature increases growth duration, thus crops can intercept more radiation \citep{muchow1990temperature}. Precipitation is an important factor. \cite{hu2003climate} found that high corn yield was associated with less rainfall in the planting period, and above-average rainfall throughout May, when seed germination and emergence happened. More rainfall with cooler temperatures were also necessary from June through August, followed by less rainfall and higher temperatures in the September–early October period. The amount of vapor pressure during growing season has impact on the variation in the potential corn yield since high vapor pressure can cause yield loss in corns \citep{zhang2017spatial}.

To evaluate the performance of the feature selection method, we obtained prediction results based on a subset of features. As such, we sorted the all features based on their estimated effects, and selected 50 most important genetic markers and 20 most important environmental components. Table \ref{tab5} shows the yield prediction performance of DNN model using these selected features. The prediction accuracy of DNN did not drop significantly compared to the corresponding results in Table \ref{tab2}, which suggested the feature selection method can successfully find the important features.

\setlength{\extrarowheight}{10pt}
\begin{table}[H]
\begin{scriptsize}
\begin{adjustwidth}{-0.0cm}{}
  \centering
 \begin{tabular}{|c|c|c|c|c|}
    \hline
{\scriptsize \begin{tabular}[t]{@{}c@{}}\\ Model\end{tabular}} &{\scriptsize \begin{tabular}[t]{@{}c@{}}Training\\RMSE\end{tabular}} &{\scriptsize \begin{tabular}[t]{@{}c@{}}Training\\Correlation\\  Coefficient (\%)\end{tabular}}& {\scriptsize \begin{tabular}[t]{@{}c@{}}Validation\\RMSE\end{tabular}}& {\scriptsize \begin{tabular}[t]{@{}c@{}}Validation\\Correlation\\  Coefficient (\%)\end{tabular}} \\
\hline
 {DNN} & 12.01& 84.01 & 12.81& 81.44   \\[7pt]
    \hline

  \end{tabular}
 \caption{Yield prediction performance of DNN on the subset of features. The DNN model used 50 genetic markers and 20 environmental components selected by feature selection method.}\label{tab5}
  \end{adjustwidth}

\end{scriptsize}
\end{table}

\section{Conclusion}

We presented a machine learning approach for crop yield prediction, which demonstrated superior performance in the 2018 Syngenta Crop Challenge using large datasets of corn hybrids. The approach used deep neural networks to make yield predictions (including yield, check yield, and yield difference) based on genotype and environment data. The carefully designed deep neural networks were able to learn nonlinear and complex relationships between genes, environmental conditions, as well as their interactions from historical data and make reasonably accurate predictions of yields for new hybrids planted in new locations with known weather conditions. Performance of the model was found to be relatively sensitive to the quality of weather prediction, which suggested the importance of weather prediction techniques.

A major limitation of the proposed model is its black box property, which is shared by many machine learning methods. Although the model captures G$\times$E interactions, its complex model structure makes it hard to produce testable hypotheses that could potentially provide biological insights. To make the model less of a black box, we performed feature selection based on the trained DNN model using backpropagation method. The feature selection approach successfully found important features, and revealed that environmental factors had a greater effect on the crop yield than genotype. Our future research is to overcome this limitation by looking for more advanced models that are not only more accurate but also more explainable.\\

\section*{Conflict of Interest Statement}

The authors declare that the research was conducted in the absence of any commercial or financial relationships that could be construed as a potential conflict of interest.

\section*{Author Contributions}

SK and LW conceived the study and wrote the paper. SK implemented the experiments.

\section*{Acknowledgments}
We thank Syngenta and the Analytics Society of INFORMS for organizing the Syngenta Crop Challenge and providing the valuable datasets. This manuscript has been released as a preprint at arXiv. The source code of the deep neural network model is available on GitHub \citep{Saeedcode}.

\section*{Data Availability Statement}
The data analyzed in this study was provided by Syngenta for 2018 Syngenta Crop Challenge. We accessed the data through annual Syngenta Crop Challenge. During the challenge, September 2017 to January 2018, the data was open to the public. Researchers who wish to access the data may do so by contacting Syngenta directly \citep{s2}.

\bibliographystyle{frontiersinSCNS_ENG_HUMS} 
\bibliography{ref}




\end{document}